\documentclass{IEEEtran}
\pdfoutput=1
\usepackage[numbers]{natbib}
\usepackage{times}
\usepackage{soul}
\usepackage{url}
\usepackage[hidelinks]{hyperref}
\usepackage[utf8]{inputenc}
\usepackage{graphicx}
\usepackage{amsmath}
\usepackage{amsthm}
\usepackage{booktabs}
\usepackage{algorithm}
\usepackage{algorithmic}

\usepackage[T1]{fontenc}

\begin{document}
\title{Mimicking Playstyle by Adapting Parameterized Behavior Trees in RTS Games}

\author{
Andrzej~Kozik,
Tomasz~Machalewski,
Mariusz~Marek,
Adrian~Ochmann
\thanks{Kozik, Machalewski, Marek and Ochmann are with Institute of Computer Science, University of Opole, Poland. Email andrzej.kozik@uni.opole.pl or tomek.machalewski@gmail.com or mariusz.marek@uni.opole.pl or aochmann@hotmail.com}
}
\maketitle

\begin{abstract}
The discovery of Behavior Trees (BTs) impacted the field of Artificial Intelligence (AI) in games, by providing flexible and natural representation of non-player characters (NPCs) logic, manageable by game-designers. Nevertheless, increased pressure on ever better NPCs AI-agents forced complexity of hand-crafted BTs to became barely-tractable and error-prone. On the other hand, while many just-launched on-line games suffer from player-shortage, the existence of AI with a~broad-range of capabilities could increase players retention. Therefore, to handle above challenges, recent trends in the field focused on automatic creation of AI-agents: from deep- and reinforcement-learning techniques to combinatorial (constrained) optimization and evolution of BTs. In this paper, we present a novel approach to semi-automatic construction of AI-agents, that mimic and generalize given human gameplays by adapting and tuning of expert-created BT under a developed similarity metric between source and BT gameplays. To this end, we formulated mixed discrete-continuous optimization problem, in which topological and functional changes of the BT are reflected in numerical variables, and constructed a dedicated hybrid-metaheuristic. The performance of presented approach was verified experimentally in a prototype real-time strategy game. Carried out experiments confirmed efficiency and perspectives of presented approach, which is going to be applied in a commercial game.
\end{abstract}


\section{Introduction}
Artificial Intelligence (AI) in computer games is attributed with great importance and responsibility - at the same time it can breathe life into an otherwise procedural, predictable and recurrent game-world, but can also make it unplayable and unnatural. Therefore, a~vast body of research has been carried out to model behaviors of Non-Player Characters (NPCs) in games, e.g.,~\cite{buckland2002ai,millington2019ai}, providing many representations and algorithms. One of them, Behavior Trees (BTs), impacted the field by providing flexible and natural representation of NPCs logic, manageable by game-designers~\cite{yannakakis, colledanchise2018behavior,florez2008dynamic}. Their success in commercial games made them implemented either as a part of game engines (CryEngine, Unreal Engine) or as plugins (Unity)~\cite{sagredo2017trained}. Nevertheless, increased pressure on ever better NPCs AI-agents forced complexity of hand-crafted BTs to became barely-tractable and error-prone, if not created by experienced AI-engineers. On the other hand, while many just-launched on-line games suffer from player-shortage, the existence of AI with a~broad-range of experience and capabilities could increase players retention~\cite{cowling2014player}. Therefore, to handle above challenges, recent trends in the field focused on automatic creation of AI-agents: from deep- and reinforcement-learning techniques~\cite{mcpartland2010reinforcement,song2019playing,liu2019playing,justesen2017learning,oh2017playing, HarmerImitation} to combinatorial (constrained) optimization and evolution of BTs~\cite{robertson2015building,tomai2014adapting,liu2016evolving,zhang2018learning}. 

Although, obtained results are impressive, they still leave room for further development and improvements. While a milestone has been reached with AlphaStar~\cite{grandmaster}, achieving a grandmaster level in StarCraft II and beating over 99\% of players, obtaining such results demand excessive resources involved in development, i.e., a lot of effort and training data.

The contribution of this paper is twofold. On the expository side we present a novel approach to semi-automatic construction of AI-agents, that mimic and generalize given human gameplays by adapting and tuning an expert-created BT, comprising a predefined options of topological and functional changes as parameterized nodes. To this end, we formulated mixed discrete-continuous optimization problem, in which parameters of the BT are reflected in numerical variables, and constructed a dedicated hybrid-metaheuristic, guided by developed similarity metric, comparing source and BT gameplays. The performance of presented approach was confirmed experimentally in a prototype Real-Time Strategy (RTS) game. 

In RTS games, the problem of developing advanced AI is particularly complicated, because of the necessity to observe a large area and to react to occurring events~\cite{RtsAI}. The AI must also simultaneously manage many units of different specifications. In addition, all decisions cannot be a mere consequence of a map situation, but must be result of a~long-term strategy, dynamically updated during a~gameplay. Therefore, our second contribution is the construction of a~complete AI-based NPC, embracing the game complexity and able to compete with human opponents.

The proposed approach is well-suited for small gamedev teams, which, with moderate effort of AI-engineers, are enabled to generate different AI-agent instances without resources needed by other methods, and, not less importantly, which are easily interpretable - by nature. In the case of considered game studio, example gameplays, defining playstyle to imitate, were directed by game-designers with use of a special tool, aiding a careful design of gameplays - with play, stop, rewind and post in-game action functionalities.

The rest of the paper is organized as follows. The next section precisely describes the rules of considered real-time strategy (RTS) game. Details of the presented approach are given in~Section~$3$, whereas in Section~$4$ settings and results of numerical experiment are presented. The last section concludes the paper.

\section{The Game}
The prototype game, provided by BAAD Games Studio, is an~RTS game, in which two players (red and green) manage their resources to find such a balance between battle and development, that either the other player is destroyed or the player acquired more resources at the end of a 15 minute gameplay. 

The game is played on a two-dimensional grid of hexagonal cells. Each cell has its $position$ on a map, and is either enabled or disabled for usage in the game - corresponding state is reflected as a background texture, e.g. water, mountains, etc.

Players manage their game entities under limited $gold$ resource. Each game entity has an unique identifier ($id$), $state$ describing its current activity (eg., $idle$, $moving$, etc.), \emph{health points} and a $class$. There are two classes of game entities - \emph{units} and \emph{buildings}. A unit represents a $quantity$ of movable forces of the same $type$, where $type$ (peasant, knight, archer) determines its combat characteristics. Buildings are able to produce resources or entities (at the expense of gold). Each building has its $type$: castle (can settle other buildings), farm (continuously delivers gold unless under attack), barracks (continuously trains assigned forces, up to its capacity limit, unless under attack), and tower (defends cells in the predefined radius by performing distanced attack on enemy units). Quantity, speed and available subset of produced entities depend on a~$type$ and a $level$ of a building.

\begin{figure}[t]
\centering
\includegraphics[width=250pt]{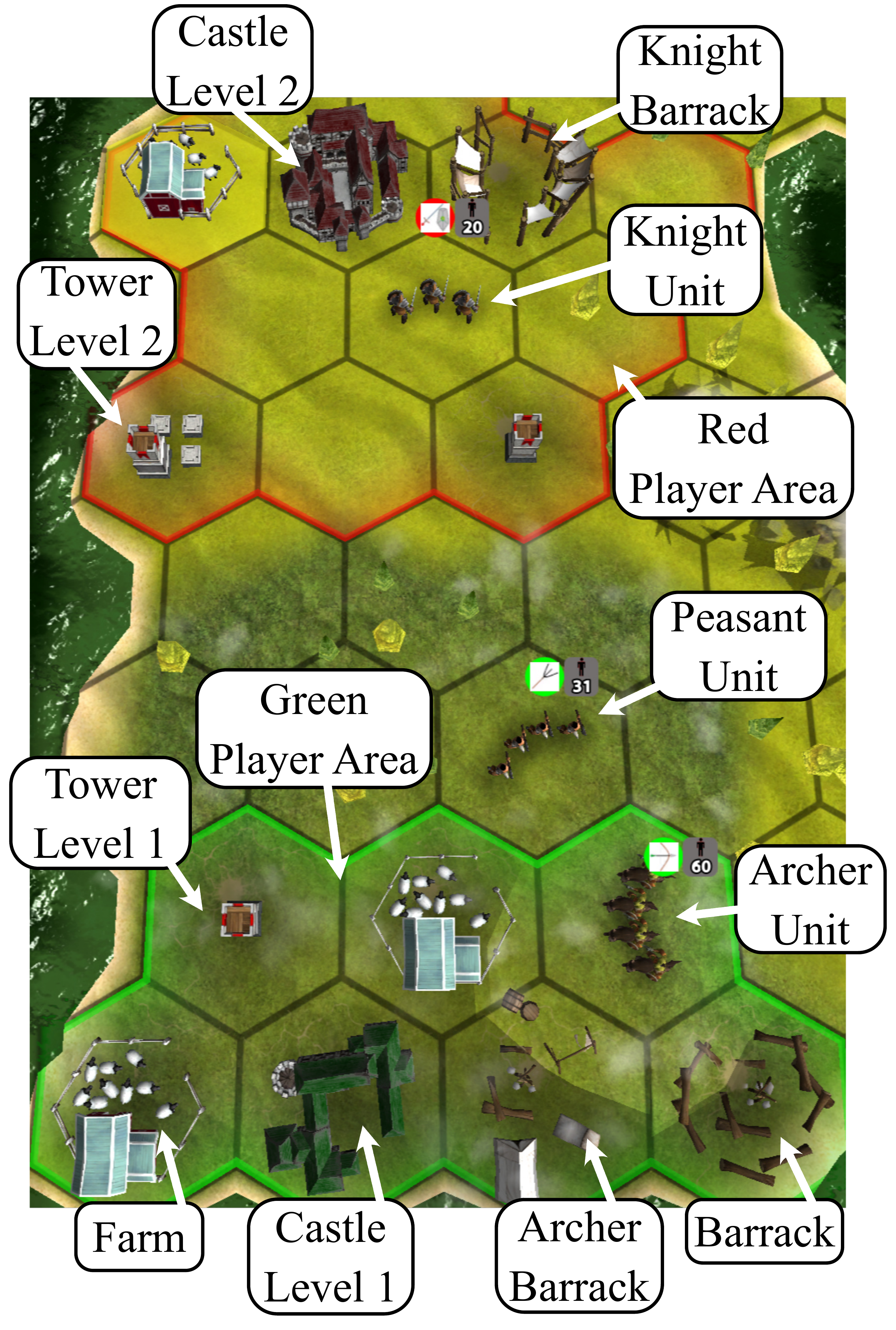}
\caption{RTS Game Scene \label{GameView}}
\end{figure}

An example scene of the game is presented in Figure~\ref{GameView}, showing the map composed of hexagonal cells. Red and green areas indicate possible building locations for each player, respectively. Different units and buildings of different levels are also presented.

During the game, all entities are controlled by issuing \emph{actions} with proper parameters. On the engine side, the game consists in $rounds$ of $1/10$ s. During each round, actions issued by each player are scheduled to be performed at the end of a~round (in the order of red-green player). 

A \emph{move} ($id$, $position$, $proportion$) action commands a~$proportion$ of the unit $id$ to move to the destination $position$, where $proportion \in \{0.25, 0.5, 0.75, 1.0\}$ (a new splitted unit is generated if $proportion < 1.0$). The unit performs its movement along the path, given by the pathfinding module, precomputed with avoidance of cells disabled or occupied by enemy entities. In the case when destination position is occupied by an unit of the same player, units are merged if both are of the same type, or the unit stops at the last feasible cell on a path, otherwise. 

In each round, every unit executes tasks, ordered by their priority: discovery of enemy unit in the attack range (same cell in the case of peasant and knight, or adjacent cell in the case of archer) and performing an attack; discovery and attack of enemy's building (while encountered on the same cell); continuation of movement along a given path; response to an attack with own attack, resulting in an uninterruptible \emph{battle}.

A \emph{spawn unit} ($id$, $type$, $quantity$) action purchases a unit of $quantity$ and $type$ from already trained in the building $id$. A \emph{settle building} ($id$, $type$, $position$) action commands the castle $id$ to settle $type$ building at $position$ on the map. \emph{Upgrade} ($id$) and \emph{repair} ($id$) actions raise the $level$ or repair of building $id$, respectively. 

The game engine provides full determinism given the same initial $seed$ value. Therefore, a gameplay can be recorded and later replayed as an \emph{Action Time-List} (ATL) for each player - an ordered collection of actions with their parameters for each round. Based on this property of the game, there are three types of players:
\begin{itemize}
\item Human-players - issuing their actions through user interface - only feasible actions are triggered in ths case.
\item ATL-players - precisely replaying a given ATL. In this case, triggering an infeasible action results in $failure$ status of the game.
\item BHT-players - AI-based players, performing actions according to a logic encoded in a Behavior Tree. A special \emph{query} action provides a data structure describing a current state of the game-world.
\end{itemize}

The prototype game is targeted to mobile market and has the complete set of features, but is limited in a variety of entity types. Note, that even such defined game poses a real challenge for AI developers to construct an algorithm guiding and managing game entities throughout the whole game.  

\section{The Methodology}

Let $\mathcal{G}(a,b)$ denote a gameplay of red ($a$) and green ($b$) players, where $a$~and $b$ are either ATLs or BTs, and let $A$ and $B$ be ATLs of a~context (source) gameplay between red and green players, respectively. The goal is to construct a BT $T$, that mimics and generalizes a playstyle of the red player, i.e., while $T$ could be autonomous AI-player,  gameplays $\mathcal{G}(A,B)$ and $\mathcal{G}(T,B)$ should be similar (according to some metric). 

Apart from similarity metric, a BT $T$ is said to be \emph{infeasible}, if its corresponding BHT-player disrupts a game reproduced by its ATL opponent, i.e., during the $\mathcal{G}(T,B)$ there is an action issued by $B$ that breaks the game with \emph{failure} status; $T$ is \emph{feasible} otherwise.

\subsection{Behavior Trees}

Classical Behavior Tree is a hierarchical structure of nodes, where each node is associated either with a task in game-world (leaf node) or performs a control-flow logic (internal node), executed in a depth-first search fashion. Execution of a node returns either $success$ or $failure$ status to its parent, depending whether its goal was achieved. This status is then used by the parent to execute or prune its remaining children.

Leaf nodes (called \emph{actions}) interact with a game-world by issuing game-actions, described in the previous section. We assume that leaves issue only a valid game-actions, and therefore always return $success$ status.

Internal tree nodes control execution flow of their children. A \emph{selector} node sequentially executes its children; if any child returns with $success$, the node stops execution and returns with $success$, otherwise it returns $failure$. A \emph{sequence} node also sequentially executes its children; the node returns with $success$ if all its children succeeded, and returns $failure$ as soon as any of them fails.

In classical BT leaf nodes called \emph{conditions} are used to check whether a given condition is satisfied in the game-world. As issuing many queries into the game may be inefficient, we developed a caching technique. Let \emph{BlackBoard} (BB) be a set of $(key, value)$ pairs, accessible to all the nodes of a BT, used as a persistent shared memory. Then, a special action node (\emph{GameQuery}) issues the \emph{query} game-action and fills BB with the current state of the game-world, to be read by the other nodes as needed.

\subsection{Adaptive Behavior Trees}
In the design of the methodology, we abandoned methods constructing BT from scratch, by iteratively evolving its topology, as they were not able to produce, in general, a reasonable BT. Note, while such a solution space can be used in searching for "the best" BT, it is still unsuitable for the considered problem, with almost all solutions evaluated as infeasible. Therefore, we adopt the approach of~\cite{ParameterizedBT}, in which a BT is pre-created by expert AI-designer in such a way, that topological and functional changes to the BT are controlled by node parameters, i.e., nodes logic can be parameterized by discrete or continuous values. Such created BT forms then a general domain to be adaptively tuned, either algorithmically or by a game-designer, to meet the expectations.

To this end, we extended the basic set of BT nodes with:
\begin{itemize}
\item time-dependent selector - executes one of its children $c_1, \ldots, c_j$ (phases) according to the current \emph{game time} and lengths of their time intervals $l_1, \ldots, l_{j-1}$, respectively.
\item switching selector - executes one of its children $c_1, \ldots, c_j$ according to the value $v \in \{1, \ldots, j\}$ of its parameter.
\item leaf-nodes with parameterized logic.
\end{itemize}

Let Adaptive Behavior Tree (ABT) $T(p)$ be a BT with parameterized nodes, where $p=\{p_1, \ldots, p_k\}$ is a vector of BT parameters, $P = P_1 \times \ldots \times P_k$ is a domain of $T(p)$ ($p \in P$) and $P_i$ is a domain of $p_i$ ($p_i \in P_i$), $i \in \{1, \ldots, k\}$. Note, all $T(p)$, $p \in P$, are feasible, i.e., each of them can be fully-functional AI-agent.

\subsection{Strategy}

\begin{figure*}[t]
\centering
\includegraphics[width=\textwidth]{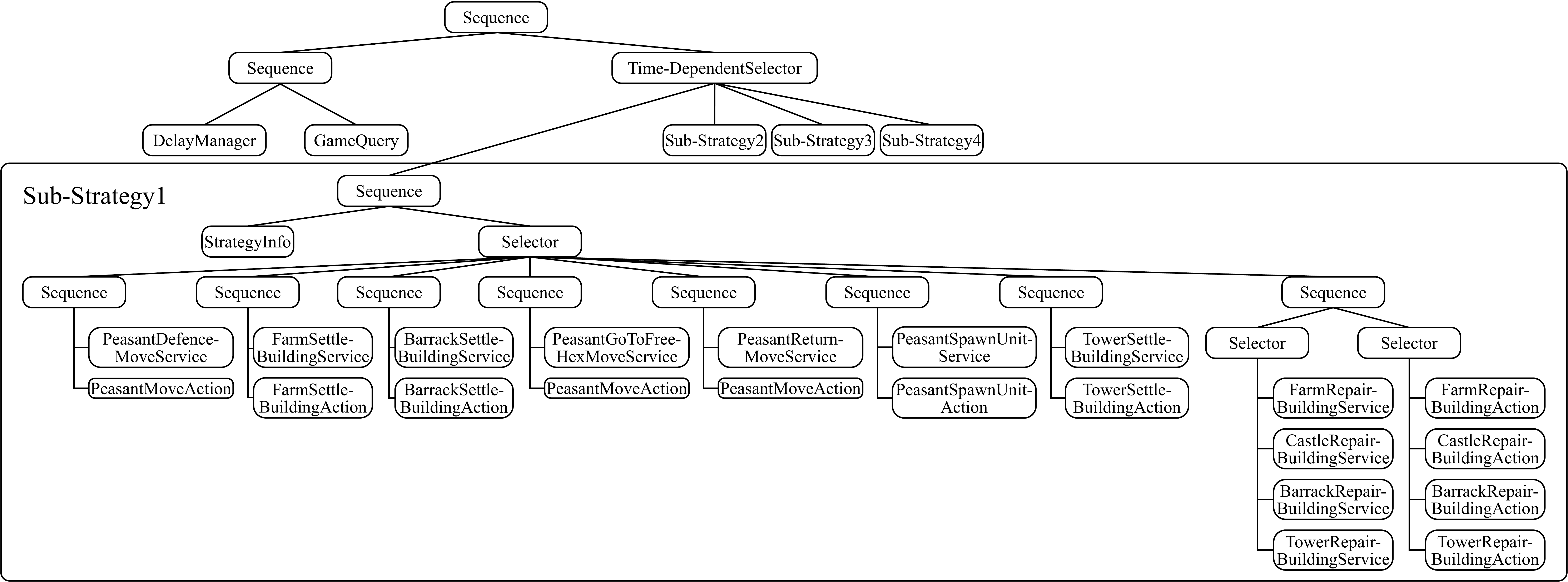}
\caption{A topology of the developed ABT. \label{SBht}}
\end{figure*}

We call an ABT a \emph{strategy}, if it can be used as an AI\nobreakdash-player, able to complete the game while competing with a player. Here, we present a strategy developed for the considered RTS game. Strategy's main idea is its ability to change the playstyle over time, due to multiple sub-strategies, i.e., initially AI can be focused on the development and defence, to become more aggressive later. Figure~\ref{SBht} shows a representative fraction of the developed ABT - core nodes and one of four sub-strategies.

At the beginning of the ABT, we perform a sequence of \emph{Black Board actions} - leaf-nodes responsible for the analysis of the game-world and providing configuration and data (as BB entries) for execution of further actions:
\begin{itemize}
	\item DelayManager – the first leaf-node in the tree, counts down the time delay between actions. If it detects (from BB) an issued game-action, it sets the delay according to type and parameters of action.
	
	\item GameQuery – a leaf-node action composed of two parts. In the first part, it reads from the game-world all available game-world state related to players:
	\begin{itemize}
		\item amount of gathered gold,
		\item list of owned buildings together with their parameters (id, position on the map, type, level, state, health-points),
		\item list of units with their properties (id, current position, type, quantity, state, health-points).
	\end{itemize}
The second part is responsible for preparation of information about the game-world, needed for creation of new buildings and units:
	\begin{itemize}
		\item empty cells available for the player,
		\item list of owned barracks buildings and currently available amounts of units ready to be bought.
	\end{itemize}
	The BB is also populated with constants defining game rules, i.e., prices and characteristics of all entities, available upgrades, list of enabled cells, etc. 
\end{itemize}
Then, we use a time-dependent selector, switching sub-strategy according to game time (total number of performed rounds).

The execution of a selected sub-strategy (e.g. Sub-Strategy~1 in Fig.~\ref{SBht}) is performed on two levels. On the first level, a general analysis of the game-world, verification of the compliance and guidance of running gameplay with the sub-strategy's goals, are performed by \emph{StrategyInfo} leaf-node. It is parameterized by a~set of tuning parameters, that configure details of its operational logic:
\begin{itemize}
	\item	desired total quantity of units,
	\item	desired total number of buildings of particular types,
	\item	preferred playstyle (offensive or defensive),
	\item	percentage balance of units between attack and defence. 
\end{itemize}
The current playstyle is also determined on the basis of present situation on the map, i.e.:
\begin{itemize}
	\item ongoing battles and their expected results,
	\item total quantity and current state of units,
	\item total number and current state of buildings.
\end{itemize}
Based on listed conditions, one of units gets selected and its task and target are determined.

The second level of sub-strategy determines the priority layout of the low-level \emph{BT services} and action nodes. BT services are divided into two types: \emph{basic services} – a simple type of nodes, whose principle of operation is to prepare the data needed to perform an assigned entity-dependent action in the game-world, and \emph{strategy services} – performing operations similar to basic services, but require direct triggering from StrategyInfo node to be successfully executed (through corresponding BB entry). Action leaf-nodes issue engine actions, corresponding to their names. Based on considered game rules and principles, the following basic services were developed:
\begin{itemize}
	\item UpgradeService – responsible for upgrading assigned buildings, depending on the received parameter.
	\item SpawnUnitService - receives 2 parameters. The first defines the minimum number of units that a building can create, the second specifies what part of the available units should be created.
	\item RepairService - finds buildings that have been damaged and then runs their repair process.
	\item GoToFreeHexMoveService - manages units that block a specific field and moves them to the nearest free field.
\end{itemize}
Similarly, the following strategy services were constructed:
\begin{itemize}
	\item SettleBuildingService - responsible for creating new buildings. As an example, a~FarmSettleBuildingService, first looks for places where it is possible to build a farm building. Then, based on game rules (the price of farm) and information about the player (owned gold and state of the castle) decides whether it is possible to build a farm in chosen location.
	\item AttackUnitMoveService - attack enemy units that count is smaller than the value specified in the parameter.
	\item AttackBuildingMoveService – receives 2 parameters. The first one defines the minimum unit group size needed to attack a building. The second parameter is the maximum life of building that will be taken into account while searching.
	\item DefenceMoveService - responsible for monitoring surroundings of the castles. If enemy units endanger the safety of the castle, they will be attacked by units responsible for its defending. The level of threat is calculated based on received parameters. The first parameter determines the size of monitored area. The second one is the minimum group size of an enemy units. As an example, a~PeasantDefenceMoveService, in case it detects an enemy on nearby cells (with detection radius and maximum quantity of enemies as node parameters), checks if StrategyInfo set (in BB) any units to defence mode and prepares parameters for subsequent engine action.
	\item UnitReturnMoveService - finds units assigned to defend a given area, that are further than a value of its parameter, and sends them back.
	\item SplitUnitMoveService - receives 3 parameters, based on which it searches for  units that exceeded the maximum allowed quantity. Then, it breaks them up according to the given proportion and sends them to a free area within the specified range. 
	\item MergeUnitMoveService - receives 3 parameters, based on which it searches for small units and merges them.
\end{itemize}

Using the described parameterized sevices and actions, we created 4 sub-strategies; given parameters values reflect capabilities of an intermediate player.

\subsubsection{Sub-strategy 1}
The aim of first sub-strategy, presented in Fig.~\ref{SBht}, is to prepare the owned territory for further sub-strategies. The goal of this state is to create at least one building of each available type. The playstyle is primarily focused on defence and development - 80\% of owned units are in defensive state. Due to the short duration of this stage, the sub-strategy does not include upgrading buildings.

\subsubsection{Sub-strategy 2}
This sub-strategy extends the previous by enabling building upgrades, creation of new unit types and building defensive towers. The strategy also includes acquisition of new territory. During this stage, 60\% of owned units are in defensive state.

\subsubsection{Sub-strategy 3}
The priorities change in this stage. About 50\% of owned units are moved to the offensive. The rest still defends the owned territory. The playstyle gets more aggressive. Offensive units attack weaker enemy units in first order and then they try to destroy buildings.

\subsubsection{Sub-strategy 4}
In the last sub-strategy, 90\% of owned units are moved to attack. Their main goal is to destroy the enemy's castles to prevent his further expansion and eventually win the game. Defence units are evenly protecting owned castles, and the observation area around castles gets increased. Grouping units into large groups becomes a priority, because of higher attack and defence strength. If insufficient units are detected during a defence battle, nearby units are searched and assigned to defend the area. 

Each of the above sub-strategies takes into account the further expansion of the owned areas, as well as maintaining the proper condition of owned buildings.

Denote the presented ABT as $T_A(p)$, $p \in P_A$. Based on similar principles we developed another strategy, $T_B(p)$, $p \in P_B$, with defence and development attitude. Note, both strategies were able to compete with their creators.

\subsection{Similarity metric}
For the purpose of measuring similarity between two gameplays, a context gameplay $\mathcal{G}_C$ and an evaluated gameplay $\mathcal{G}_E$, we developed a metric, which heuristically assesses similarity between two multivariate time series, representing a generalized views on corresponding gameplays, from the evaluated player perspective.

Let $S_\mathcal{G} = (S_\mathcal{G}^1, \ldots, S_\mathcal{G}^n )$ be a Snapshot Time-Line of a~gameplay $\mathcal{G}$, where $S_\mathcal{G}^i = (S_\mathcal{G}^{i,1}, \ldots,    S_\mathcal{G}^{i,k})$ (a game snapshot at time $i$) describes a state of player's resources $S_\mathcal{G}^{i,1}, \ldots,    S_\mathcal{G}^{i,k}$ at time $i$, $i \in \{1, \ldots, n\}$ and $n$ is a number of snapshots, assuming they are sampled with a constant interval throughout the gameplay. In the presented approach, we sampled the amount of gold, the total number of units, the total number of buildings and numbers of entities of the same type as a representation of a game-state. 

Let $S_{\mathcal{G}_C}$ and $S_{\mathcal{G}_E}$ be min-max normalized to $[-1, 1]$ interval. Then, a similarity matrix $d$, inspired by Self-Similarity matrix, often used recently and in the past in sequential data processing research~\cite{hirai2019melody2vec, Jun2013MusicSA}, is computed in such a~way, that each element $d_{i,j}$ represents similarity between $S_{\mathcal{G}_C}^i$ and $S_{\mathcal{G}_E}^j$ computed as averaged Manhattan distance, i.e.,

\begin{equation*}
d_{i,j} = \frac{1}{k} {\| S_{\mathcal{G}_C}^i, S_{\mathcal{G}_E}^j \|}_1 = \frac{1}{k} \sum_{l=1}^{k}|S_{\mathcal{G}_C}^{i,l}-S_{\mathcal{G}_E}^{j,l}|.
\end{equation*}

\begin{figure}[t]
\centering
\includegraphics[width=240pt]{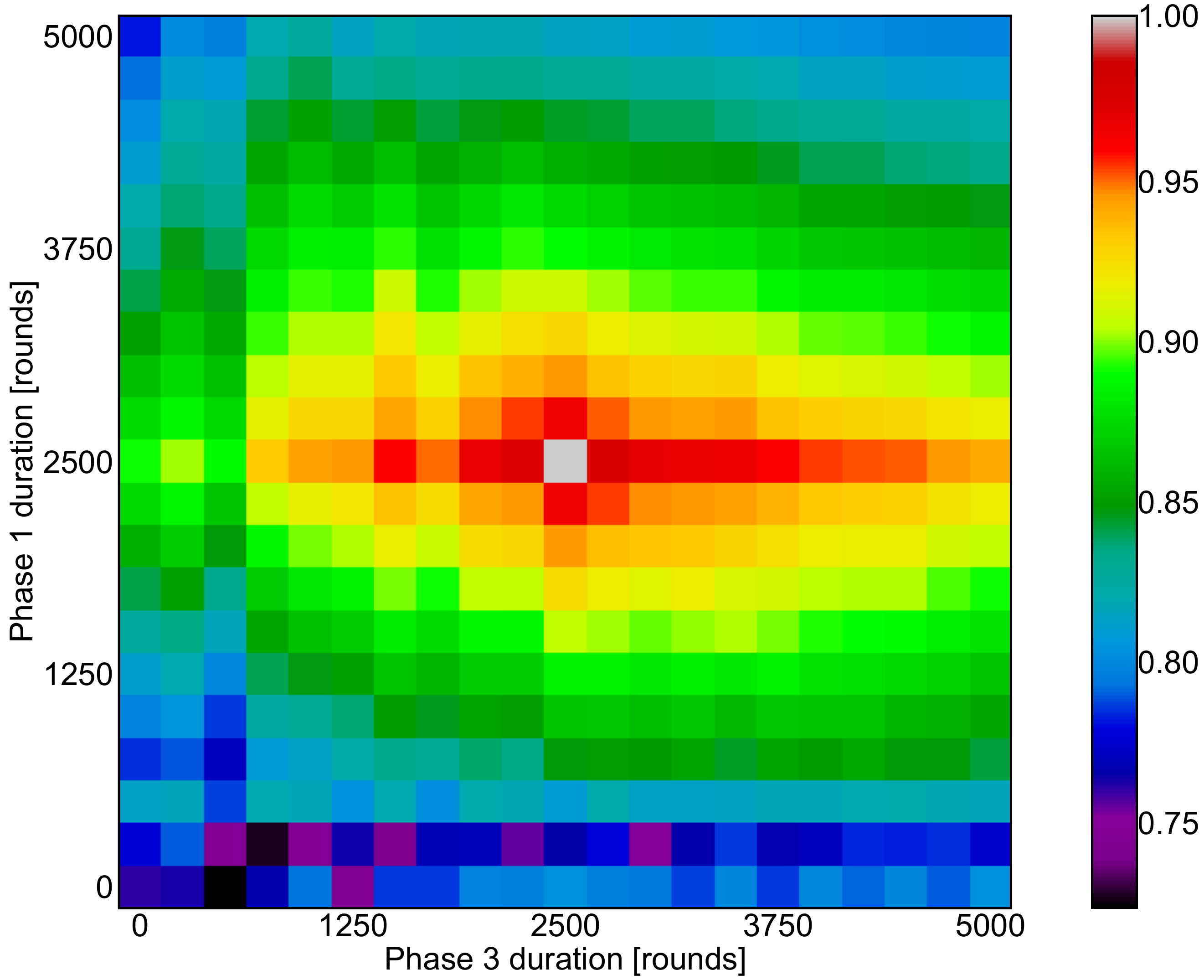}
\caption{Similarity metric values for different parameters of ABT.  \label{Gradient}}
\end{figure}

The main idea of the presented approach is to promote similarities, in which pairs of non-distant snapshots are appearing after each other in the matrix $d$. To this end, let a trend $t$ be a series of diagonal elements of matrix $d$, i.e., a trend $t$ starting on $d_{i,j}$ of length $\|t\| = l$ is defined as 
\begin{equation*}
t = (d_{i,j}, d_{i+1,j+1}, \ldots, d_{i+l,j+l}).
\end{equation*}
Let $v(t)$, an evaluation of $t$, equal
\begin{equation*}
v(t) =(\sum_{x\in t}x)*\|t\|^2.
\end{equation*}
We defined a trend-series $\mathcal{T} = \{t_1, \ldots, t_z\}$ as a set of trends, for which the following conditions hold:
\begin{itemize}
\item there is exactly one cell in each column of $d$ belonging to some trend in $\mathcal{T}$,
\item no two trends from $\mathcal{T}$ can be merged to form a longer trend.
\end{itemize}
Let an evaluation of $\mathcal{T}$, denoted as $V(\mathcal{T})$, equal
\begin{equation*}
V(\mathcal{T}) = \frac{\sum_{t\in \mathcal{T}}v(t)}{(\sum_{t\in \mathcal{T}}\|t\|)^3}.
\end{equation*}
Then, a similarity metric $\mathcal{S}\Big(\mathcal{G}_C, \mathcal{G}_E \Big)$ between $\mathcal{G}_C$ and $\mathcal{G}_E$ gameplays is equal to a maximum value of trend-series over similarity matrix $d$ obtained from $\mathcal{G}_C$ and $\mathcal{G}_E$. Corresponding value can be computed by a simple algorithm, utilizing prefix-sums and dynamic programming, of time complexity $O(\max(m, n)^3)$, where $m$ and $n$ are length of two Snapshot Time-Lines.

An example application of developed metric is presented in Fig.~\ref{Gradient}, showing values of similarity between a context game of $T_A$ and $T_B$ with fixed parameters, and gameplays in which values of two $T_A$ parameters were changed, i.e., for the context gameplay the lengths of the first and third phases in the time-dependent selector node was set to 2500 rounds, while for evaluated gameplays their value was changed from 0 to 5000 rounds.

A distinct gradient can be seen, with peak reached at the point $(2500, 2500)$, reflecting comparison with target match. Similarity greatly differs in the vertical axis, the length of first phase - similarity metric appropriately distinguishes games varying in this parameter, because length of this phase strongly affects later course of the game. The metric has more difficulty in differentiating games varying in length of the third phase, that is justified by its similarity with the fourth phase, which complements the remaining game, up to 15 minutes (10000 rounds).

A similarity landscape, generated by developed metric, exhibits properties desirable by gradient-based optimization techniques. It properly reflects continuous changes in parameters values in the neighborhood of the global optimum - around values of context gameplay. On the other hand, discontinuities and small local extrema are to be dealt with design of metaheuristic optimization algorithm.

\subsection{Optimization Problem}
Given an ABT $T(p)$, its domain $P$ and a context ATLs $A$~and $B$, the goal is to find such $p^* \in P$, that $p^*$ is feasible and similarity metric $\mathcal{S}$ between $\mathcal{G}(A, B)$ and $\mathcal{G}(T(p^*), B)$ is maximized, i.e.,
\begin{equation}
\label{OPP}
p^* = \underset{p \in P'}{\mathrm{arg\,max}} {\mathcal{S}\Big(\mathcal{G}(A, B), \mathcal{G}(T(p), B)\Big)},
\end{equation}
where $P' = \{p \in P | T(p)\textrm{ is feasible}\}$ ($\mathcal{G}(T(p), B)$ does not end with a~\emph{failure} status). 

\section{Experimental evaluation}
 
The methodology developed in Section 3 allows to cast a~problem of automatic construction of AI-agents, that mimic and generalize given human gameplay, as an optimization problem. Let $P$, a domain of parameters of a given ABT, be a solution space to be searched for a solution $p^* \in P$ solving (\ref{OPP}). 
Such an optimization problem is characterized by a nontrivial objective function, requiring simulation of a gameplay for each solution $p \in P$.

To solve formulated optimization problem, in this section a~hybrid metaheuristic~\cite{blum2008hybrid} based on Memetic Search~\cite{MemeticSearch} is developed, to verify applicability and performance of the presented approach. Note, the presented algorithm is not constructed with time efficiency in mind. It is a proof of concept and a hint of promising optimization techniques, since in a solution to be deployed, the optimization problem is solved by a cloud-based parallel algorithm, tackling many Behaviour Trees at once with hierarchical optimization techniques, and using a scalable pool of game simulators.
 
\subsection{Metaheuristic optimization algorithm}
As a solution search environment we constructed an algorithm based on the Memetic Search hybrid-metaheuristic, which combines the strength of evolutionary algorithms in diversification of the search space exploration with complementary search intensification property of driven trajectory-method.  

\subsubsection{Criterion} 
The solution search process is obviously driven by the similarity metric $\mathcal{S}$ between context and evaluated gameplays, defined in Section~3.4. Nevertheless, not all solutions from $P$ are feasible. In this case, we introduced a \emph{penalty} for infeasible solutions:
\begin{equation*}
penalty = \Big|1 - \frac{\|\mathcal{G}_E\|}{\|\mathcal{G}_C\|}\Big|,
\end{equation*}
where $\mathcal{G}_C$ is a context gameplay, $\mathcal{G}_E$ is an evaluated gameplay and $\|\mathcal{G}\|$ denotes the length of a gameplay $\mathcal{G}$ (the number of rounds). Note, the less rounds were performed in an evaluated gameplay the greater a $penalty$ value.

Finally, the criterion value $f(p)$ of a solution $p$ used in the presented algorithm equals
{\setlength\arraycolsep{2pt}
\begin{eqnarray}
f(p) & = & \mathcal{S}\Big(\mathcal{G}(A, B), \mathcal{G}(T(p), B)\Big)-{}
\nonumber\\
& & \Big|1 - \frac{\|\mathcal{G}(T(p), B)\|}{\|\mathcal{G}(A, B)\|}\Big|, \label{Criterion}
\end{eqnarray}}
where $A$ and $B$ are ATLs of a~context gameplay and $T(p)$ is evaluated ABT.

Since computation of a similarity measure between context and evaluated gameplays is a computationally demanding task, we implemented the algorithm in such a way, that once a solution is evaluated, its corresponding criterion value is cached for future usage.

\subsubsection{Memetic Search}
The Memetic Algorithm~\cite{MemeticSearch} in its basic form is a classic evolutionary algorithm, iteratively managing a set of solutions (a population) by applying crossover and mutation operators, additionally using an improvement algorithm as an intensification strategy.

\begin{figure}[t]
\centering
\includegraphics[width=240pt]{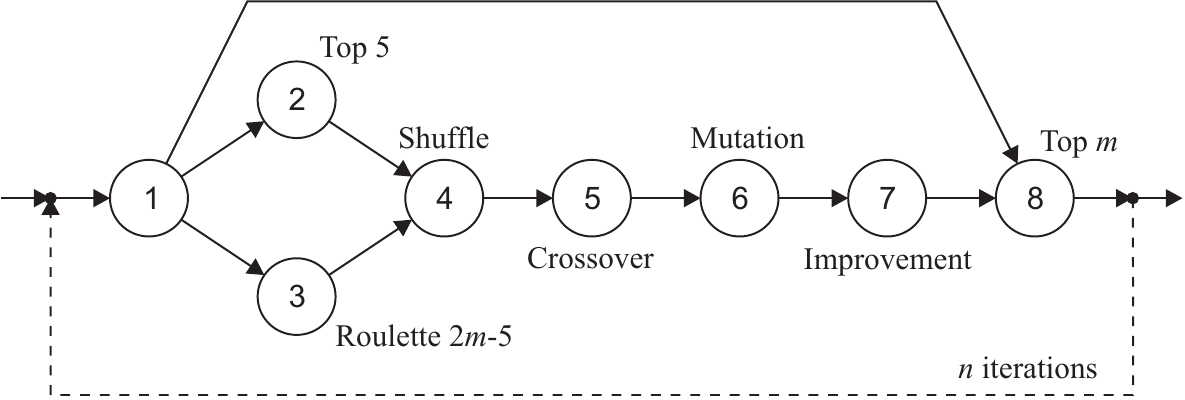}
\caption{The Memetic Search algorithm. \label{Memetic}}
\end{figure}

To model and implement the algorithm we developed a~\emph{Heuristics Composition Engine}, in which algorithms are modeled as directed acyclic graphs. Nodes of such graphs represent some tasks performed on populations of solutions, and edges represent control and population flows. Nodes are executed in a topological order. An executed node forms its input population by merging output populations of its predecessors in a graph, and then computes its output population by performing its task. A graph corresponding to developed Memetic Search algorithm is presented in Figure~\ref{Memetic}.

Consider an initial population consisting of $m$ solutions. The first node has an empty task - it only transfers initial population to subsequent nodes. In node $4$ an input population, created from the best~$5$ solutions from initial population (node~$2$) and from $2m-5$ solutions selected using \emph{roulette wheel} rule (node~$3$), is shuffled. In node~$5$ a crossover operator is applied to consecutive pairs of solutions. Recombined solutions are mutated (with a small probability) in node~$6$. Node~$7$ applies an improvement algorithm to each solution from the input population, and finally node~$8$ selects the best $m$ solutions from initial and improved populations. The procedure is repeated for $n$ iterations.

\subsubsection{Intensification strategy}
As an intensification strategy we used a Simulated Annealing (SA)~\cite{SimulatedAnnealing}. The SA starts the search process from a given initial solution $p$. Then, at each iteration a new solution $p'$ is randomly sampled from a neighborhood.The new solution $p'$ replaces the old solution $p$ either with a probability computed following the Boltzmann distribution $e^{-\frac{f(p)-f(p')}{\tau}}$ if $f(p') \leq f(p)$, or without a draw if $f(p') > f(p)$. The so-called temperature $\tau$ is decreased after each iteration $i=1,2,\ldots,i_{max}$ by a~geometric cooling schedule, i.e., $\tau_i = \alpha \tau_{i-1}$, $\alpha \in (0,1)$. 

\subsection{Numerical Experiment}
Taking into account limits induced by utilized delay-manager BT node, to show applicability  of the presented approach, we designed a synthetic experiment setup as follows.

A gameplay $\mathcal{G}(T_A(p_A), T_B(p_B))$ of two ABTs $T_A(p)$ and $T_B(p)$ with fixed parameters $p_A \in P_A$ and $p_B \in P_B$ was recorded as a~context gameplay. The ABTs represented two different strategies, described in Section 3.3.

The goal of the Memetic Search algorithm is to (re)discover of the parameters of a BHT-player. In the experiment, we use three ABT-driven players:
\begin{itemize}
\item Player 1 - $T_A(p)$, $p \in P_A$ - the ABT from the context gameplay,
\item Player 2 - $T_A(p)$, $p \in P'_A$ - the domain $P'_A$ of $T_A(p)$ was limited for one parameter in such a way, that $p_A \notin P'_A$,
\item Player 3 - $T_B(p)$, $p \in P_B$ - the ABT of the opponent in the context gameplay.
\end{itemize}
As a by-product of such a setup, we eliminated the need for estimating delays between actions performed by human players, as in both context and evaluated gameplays delays are managed in the same way. On the other hand, since the constructed ABTs limit the rate of performed actions to at most one per second, the frequency of changes observed in the game-world is about $1$~Hz. Therefore, based on the Sampling Theorem~\cite{Sampling}, we set the sampling frequency of game snapshots to be 2~Hz\footnote{In a performed experiment (not showed here) we obtain that 4~Hz sampling rate had no effect on the convergence of the presented algorithm, while it dramatically increased its run-time, due to $O(n^3)$ complexity of similarity metric. On the other hand, with~1~Hz sampling frequency the algorithm was not able to find a good solution.}.

Algorithms were coded in C\# and simulations were run on a PC with CPU Intel Core i7-3610QM 2.30~GHz and 16GB RAM. Parameters of algorithms were chosen empirically as follows: for the Memetic Search algorithm we set $m = 12$ and $n = 20$, and for subordinate SA we set $\tau_0 = 50$, $\alpha = 0.998$ and $i_{max} = 5$. The initial population was drawn randomly from the domains of corresponding ABTs. Due to non-deterministic nature of developed algorithms, each was ran $100$ times, yielding 1 hour per run on the average. 

\begin{figure}[t]
\centering
\includegraphics[width=240pt]{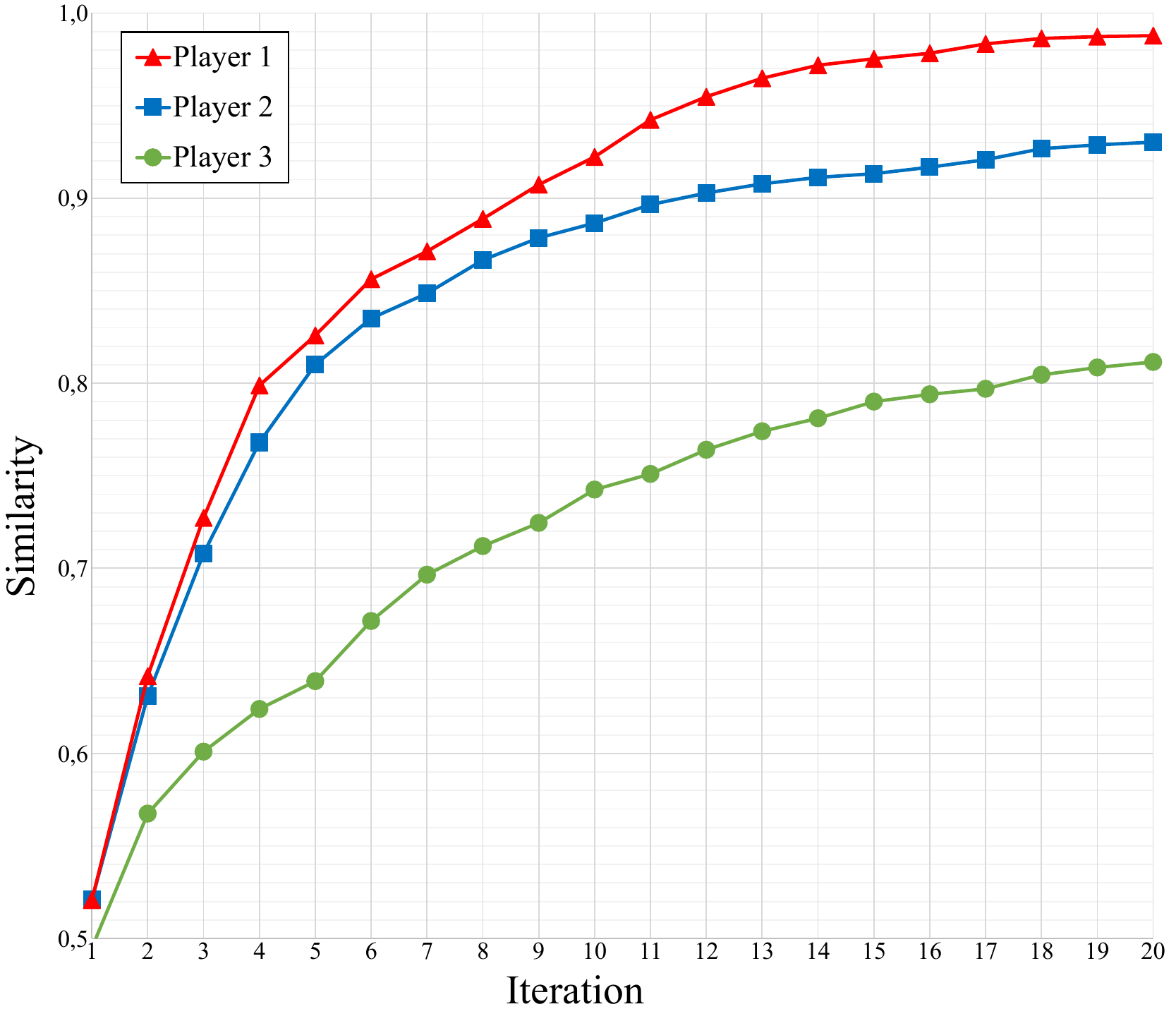}
\caption{Results of numerical experiment. \label{Wykres}}
\end{figure}

Figure~\ref{Wykres} presents average values of similarity metric for the best found solution in subsequent iterations of the Memetic Search algorithm. It can be seen, that developed heuristic similarity metric properly assess likelihood and differences between gameplays - parameters of Player~1 was determined correctly, reaching almost perfect similarity, for Player~2 the similarity saturated at the smaller value ($0.93$) - but reached limitations of its domain, and the value obtained for Player~3 clearly renders his playstyle dissimilar.

\section{Conclusions}

In this paper, we presented a novel approach to automatic construction of AI-agents, that mimic and generalize given human gameplays by adapting and tuning of ABTs - parameterized BTs, characterized by varying behaviors, diverse strategies and a range of skills and capabilities. To this end, we formulated mixed discrete-continuous optimization problem, in which topological and functional changes of the BT are reflected in numerical variables, and constructed a~dedicated hybrid-metaheuristic, driven by developed similarity metric between source and BT gameplays. The performance of presented approach was confirmed experimentally on a prototype RTS game - ABT can be tuned so that it mimics human gameplay, given that it covers his playstyle.

The future work will be concentrated on mathematical models of delays in human gameplays~\cite{delays}. In the presented approach, evaluated trees share the same simple model, therefore, it does not impact achievable similarity between parameterized BTs gameplays. On the other hand, adequate human-delay model, together with growing set of ABTs, covering a wide repertoire of playstyles, will enable on-demand generation of "ghost-players" - AI-agents mimicking requested human opponents, and in turn, monetization of such a feature. 

\section*{Acknowledgments}
The work was financially supported by the National Centre of Research and Development in Poland within GameINN programme under grant no. POIR.01.02.00-00-0108/16.

\end{document}